%% file: template.tex
\documentclass{article}

\usepackage{arxiv}

\usepackage{threeparttable}
\usepackage[utf8]{inputenc} 
\usepackage[T1]{fontenc}    
\usepackage{hyperref}       
\usepackage{url}            
\usepackage{booktabs}       
\usepackage{amsfonts}       
\usepackage{nicefrac}       
\usepackage{microtype}      
\usepackage{lipsum}
\usepackage{graphicx}
\usepackage[authoryear]{natbib}

\graphicspath{ {./images/} }
\usepackage{newtxtext,newtxmath}  
\usepackage{natbib}
\usepackage{array} 
\usepackage{tabularx}
\usepackage{enumitem}
\usepackage[table]{xcolor}
\definecolor{lightgray}{gray}{0.9}
\usepackage{geometry}
\usepackage{longtable}
\usepackage{caption}
\usepackage{svg}
\usepackage[multiple]{footmisc}
\usepackage{url}
\usepackage{amsmath}
\usepackage{pifont} 
\newcommand{\xmark}{\ding{55}} 
\usepackage{multirow}
\newcommand{\cmark}{\ding{51}} 
\usepackage{subcaption}
\usepackage{hyperref}
\usepackage{threeparttable}
\usepackage{makecell} 

\usepackage{pifont}    
\usepackage{xcolor}    

\newcommand{\myxmark}{{\color{gray!40}\xmark{}}}

\title{Food for thought: How can machine learning help better predict and understand changes in food prices?}

\author{
  Kristina L. Kupferschmidt \\
  School of Engineering, University of Guelph, Guelph, Canada\\
  Vector Institute, Toronto, Canada\\
  \texttt{kupfersk@uoguelph.ca} \\
  \And
  James Requiema \\
  University of Toronto, Toronto, Ontario, Canada\\
  Vector Institute, Toronto, Canada\\
  \texttt{requeima@cs.toronto.edu} \\
  \And
  Mya Simpson\thanks{Equal authorship.} \\
  School of Engineering, University of Guelph, Guelph, Canada\\
  \texttt{msimps07@uoguelph.ca} \\
  \And
  Zohrah Varsally\footnotemark[1] \\
  School of Engineering, University of Guelph, Guelph, Canada\\
  \texttt{varsallz@uoguelph.ca} \\
  \And
  Ethan Jackson \\
  \texttt{jackson.ethan.c@gmail.com} \\
  \And
  Cody Kupferschmidt \\
  \texttt{cody@erode.ai} \\
  \And
  Sara El-Shawa\\
  \texttt{saraelshawa1@gmail.com} 
  \And
  Graham W. Taylor\\
  School of Engineering, University of Guelph, Guelph, Canada\\
  Vector Institute, Toronto, Canada\\
  \texttt{gwtaylor@uoguelph.ca} \\
  }

\begin{document}
\maketitle
\begin{abstract}
In this work, we address a lack of systematic understanding of fluctuations in food affordability in Canada. Canada's Food Price Report (CPFR) is an annual publication that predicts food inflation over the next calendar year. The published predictions are a collaborative effort between forecasting teams that each employ their own approach at Canadian Universities: Dalhousie University, the University of British Columbia, the University of Saskatchewan, and the University of Guelph/Vector Institute. While the University of Guelph/Vector Institute forecasting team has leveraged machine learning (ML) in previous reports, the most recent editions (2024--2025) have also included a human-in-the-loop approach. For the 2025 report, this focus was expanded to evaluate several different data-centric approaches to improve forecast accuracy. In this study, we evaluate how different types of forecasting models perform when estimating food price fluctuations. We also examine the sensitivity of models that curate time series data representing key factors in food pricing.
\end{abstract}


\section{Introduction}
Food affordability is a prominent issue in Canada, with many Canadians struggling to cope with recent increases in the cost of living. The price of food is known to be impacted by many factors including climate events, supply chain disruptions, policy changes (e.g. carbon taxing), and other geopolitical factors \citep{charlebois2024digital, charlebois2024implications, Kalkuhl2016-xc}. In 2023, affordability concerns were particularly pronounced due to high inflation, which was accompanied by downstream increases in the costs of food, housing, energy, and other expenditures \citep{Kryvtsov2023-xm}. Although many researchers have attempted to quantify the impacts of specific factors on food prices, the main factors are not independent, and interactions between multiple regressors must be considered \citep{Kalkuhl2016-xc}.

Canada's Food Price Report (CFPR)\footnote{\url{https://www.dal.ca/sites/agri-food/research/canada-s-food-price-report-2024.html}} has been published for 14 years, and provides expectations of annual food expenditures for average Canadians. The 2025 CFPR was produced through a Canada-wide collaboration between Dalhousie University, the University of Guelph/Vector Institute, the University of British Columbia, and the University of Saskatchewan. The CFPR has two main objectives: (1) to help Canadians better understand and create budgets to allow them to continue buying the necessary food for themselves and their dependents and (2) to provide policymakers with an estimate of food price inflation. Every autumn, the CFPR focuses on predicting the percentage change for 9 food categories in the Consumer Price Index (CPI) from Statistics Canada\footnote{\url{https://www150.statcan.gc.ca/t1/tbl1/en/tv.action?pid=1810000401}} for the upcoming year. The 2025 CFPR is scheduled to be released on December 5, 2024.

The CFPR employs a unique hybrid approach, combining judgmental forecasts with other statistical and ML-based approaches. Teams at the associated universities work together to implement and present different approaches to forecasting anticipated changes to food prices. These forecasts are then combined by a diverse team of experts in agriculture, management, economics, and computer science to produce a single predicted range for each CPI food category. The 2025 forecasts created by the University of Guelph/Vector Institute team focused on following a data-centric approach guided by human-centered design practices. This technical report complements Canada's Food Price Report by providing more details about our team's approach to forecasting.

In the early stages of problem definition, domain experts were consulted to identify important factors driving food prices and provide suggestions on where to find and how to incorporate these data. These suggestions were then used to complete an extensive web scraping exercise, where regressor time series that reflected the expert-identified factors were selected and pre-processed. A sensitivity analysis was conducted to understand the impacts of data curation using methods such as thematic grouping of regressors and the application of large language models (LLMs) to curate bespoke sets of regressors for specific tasks. 

\subsection{Objectives}
While the primary aim of the aforementioned work contributing to the 2025 Canada's Food Price Report was to produce high-quality forecasts while minimizing error over the selected test periods, several additional research questions emerged. Specifically, the objectives of this work were four-fold:

\begin{enumerate}[label=\arabic*.]
    \item Adapt an existing human-centric forecasting framework to use a combination of human and machine experts to maximize forecasting performance.
    \item Explore methods for incorporating additional context into time-series forecasting to improve accuracy.
    \item Investigate whether intrinsic characteristics of time-series data (e.g., food categories) influence the effectiveness of high-capacity ML models.
    \item Compare different model and data curation methods included in the forecasting experiments conducted for the 2025 CFPR.
\end{enumerate}

\section{Related Works}
\subsection{Human-centric forecasting}
ML-based forecasting continues to demonstrate gradual advancements, however, there remain opportunities to improve alignment in the delivery of these predictions. Without involving end-users, there is a risk that proposed methodologies will fail to generalize to future forecasting tasks and user needs there were not previously identified may emerge. To address these concerns, human-centered frameworks, such as DelphAI, were designed to facilitate expert engagement throughout the development pipeline to guide decisions to improve forecasting accuracy and usefulness \citep{kupferschmidt}.  While DelphAI remains domain and task-agnostic, we have identified two potential ways of adapting the framework to address the goals of the CFPR: (1) consulting experts to identify potential exogenous data sources understood to be related to the target task, and (2) engaging these experts to actually generate time-series forecasts. 

\subsection{Incorporating context into forecasting}

\begin{figure}[h]
    \centering
    \includegraphics[width=0.5\linewidth]{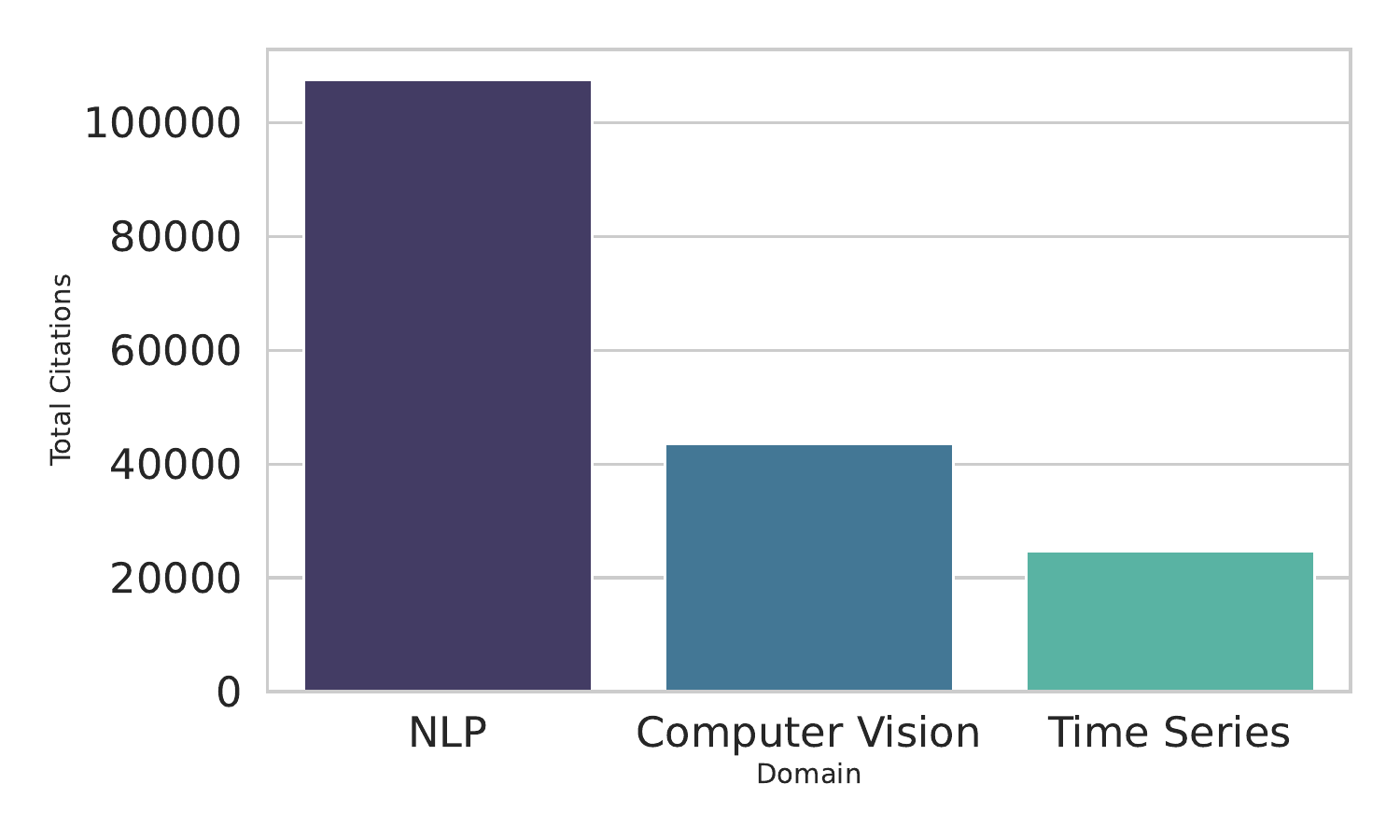}
    \caption{Comparison of citation trends across different domains of foundation models.\protect\footnotemark}
    \label{fig:citation-comparison}
\end{figure}

In recent years we have seen the rapid adoption of foundation models, which are general-purpose and use self-supervised learning at scale to complete tasks across different domains. Several prominent time-series foundation models, such as TimeGPT \citep{garza2023timegpt1} and Chronos \citep{ansari2024chronos}, have achieved state-of-the-art results in zero-shot forecasting. However, despite growing interest in the use of foundation models for time series, adoption has been slow compared to the incredible uptake of foundation models in fields such as natural language processing (NLP) and computer vision (Figure \ref{fig:citation-comparison}). This discrepancy is often attributed to the limited contextual information provided to time-series forecasting models, including a lack of details about the task and the relationships between other provided features.

\footnotetext{The data includes keyword searches for the following domains: 
\textbf{NLP:} "Foundation models NLP", "Foundation models language", "Transformer language models", "Large language models"; 
\textbf{Computer Vision:} "Foundation models computer vision", "Foundation models vision", "Vision Transformers"; 
\textbf{Time Series:} "Foundation models time series", "Foundation models forecasting", "Transformers time series", "Transformers forecasting".}

There have been recent studies that have demonstrated the value of using LLMs for time-series forecasting \citep{jin2023large, gruver2024large}. Notably, several works have explored new methodologies for providing LLMs with different types of context to complete forecasting tasks, with the goal of allowing LLMs to leverage the underlying knowledge on the subject or related subjects that may have been experienced during pre-training {\citep{requeima2024llm, williams2024context}. \citet{requeima2024llm} proposed \textit{LLM Processes}, a technique where LLMs can be repeatedly probed to create regression (and time-series) predictions. Building upon this work, \citet{williams2024context} proposed \textit{Context is Key}, and more specifically the \textit{Direct Prompt} framework, where a model is directly instructed to forecast a structured output rather than prompt it repeatedly. 

\section{Methodology}
A new dataset was aggregated from publicly available sources such as Statistics Canada \footnote{\url{https://www150.statcan.gc.ca}}, the US Federal Reserve Dataset (FRED) \footnote{\url{https://fred.stlouisfed.org/}}, and the US National Oceanic and Atmospheric Administration \footnote{\url{https://www.noaa.gov/}}. These data were included as either (1) target variables that are being predicted by the forecasting model, or (2) external regressors to provide additional context to the model when predicting the target variable. 

\subsection{Target task and dataset --  Predicting food-related consumer price index}
Statistics Canada publishes a Consumer Price Index (CPI), which serves as a comprehensive indicator of inflation and the cost of living in Canada. Specifically, with respect to food, the CPI provides pricing information for nine distinct food categories:
\begin{enumerate}[label=\arabic*.]
    \item Bakery and cereal products (excluding baby food)
    \item Dairy products and eggs
    \item Fish, seafood, and other marine products
    \item Food purchased from restaurants
    \item Fruit, fruit preparations, and nuts
    \item Meat
    \item Other food products and non-alcoholic beverages
    \item Vegetables and vegetable preparations
    \item Food (general across all other categories)
\end{enumerate}

Historical monthly CPI data for each food category from 1986 to 2017 were used to train the individual models. Given the annual periodic nature of the CFPR, the models were evaluated based on their ability to produce 18-month forecasts for the six most recent years (2018 to 2024). For all models, the context length remained consistent at 36 months with the exception of the LLMs which had a context window of 75 months to provide longer historical look back period. Models were evaluated for their mean absolute percentage error (MAPE) (Equation \ref{eq:mape}) predicting a single food CPI category at a time. Performance metrics were calculated using MAPE scores between the forecasted values and ground truth across the entire test period. 

\begin{equation}
\label{eq:mape}
\text{MAPE} = \frac{1}{n} \sum_{i=1}^{n} \left| \frac{y_i - \hat{y}_i}{y_i} \right| \times 100
\end{equation}

\subsection{Data curation to provide specific context}
Through web scraping, a collection of time-series regressors was created to represent food price drivers across the food production chain, with the hypothesis that forecasting models could be improved through additional context. This process resulted in the collection of 165 unique time series.

\subsubsection{Domain expert data curation}
Predicting future food prices is known to be a complex task, dependent on many frequently shifting factors. For the 2024 and 2025 CFPR, collaborators with extensive experience in food-driven economics were consulted to identify regressors that they expected would have the greatest impact on Canadian food prices. Several key themes emerged from the expert-generated suggestions: economic, climate, geopolitical, and manufacturing factors ( Appendix \ref{appendix:groupings of variables}).
\\

\textit{\textbf{Theme 1) Economic factors}}: Data from three general economic categories were grouped: goods prices (spot crude oil prices, U.S. imports of goods, electricity prices, and CAD/USD exchange rate, Energy CPI), employment data (number of employees and active unemployment rate in the United States, number of work stoppages), and price indices (interest rates, Federal Funds Effective rate, producer price indices for railroad, pesticides, and fertilizer, and US Recession probability rate). These data were collected from the US Federal Reserve Database \footnote{\url{https://fred.stlouisfed.org/}} and Statistics Canada\footnote{\url{https://www150.statcan.gc.ca/t1/tbl1/en/tv.action?pid=1810000401}}. 

\textit{\textbf{Theme 2) Climate factors}:} Climate data was selected based on expected impacts on food production. Three proxy data sources were used to represent drought and precipitation conditions for common growing regions across North America: ENSO index (a measure of the strength of the El-Niño Southern Oscillation\footnote{\url{https://psl.noaa.gov/enso/mei/}}, which is known to affect precipitation patterns globally), Palmer Drought Severity Index (a measure of relative dryness, filtered for key agricultural regions and states within the United States)\footnote{\url{https://www.ncei.noaa.gov/pub/data/cirs/climdiv/climdiv-pdsist-v1.0.0-20240705}}, and California snowpack (water stored as snow in agricultural regions of California, where snowmelt is a major source of irrigation)\footnote{\url{https://cdec.water.ca.gov/dynamicapp/wsSensorData}}.

\textit{\textbf{Theme 3) Geopolitical factors}:} Proxy data was selected to reflect that geopolitical instability, such as the Russian invasion of Ukraine, which has been a major driver of food price increases in recent years, particularly within the Bakery category \citep{Kalkuhl2016-xc, carter2024did}. Although many datasets were initially identified very few were available for the entire historical training period beginning in 1986. Data included the Canada News-Based Policy Uncertainty Index\footnote{\url{https://www.policyuncertainty.com/canada_monthly.html}}, FRED currency exchange rates, customs rates of imports separated by major trade countries, and global commodity price indices from the World Bank\footnote{\url{https://www.worldbank.org/en/research/commodity-markets}}.

\textit{\textbf{Theme 4) Manufacturing factors}: }Data to reflect the state of manufacturing were aggregated primarily from FRED. In particular, these data sources included food manufacturing and producer price indices separated by sector as well as signals related to transportation.

\subsubsection{LLM-persona data curation}
In addition to the human-curated list by themes, LLMs were also evaluated for their ability to complete data curation tasks. GPT-4o was prompted to help rate how valuable it thought different external regressors would be for forecasting Canadian food prices. In each prompt, the LLM was instructed to assume the knowledge base of a particular persona \citep{schuller2024generating} and was provided with a layperson description of each external variable.  The construction of personas was motivated through the principles ingrained in popular judgmental forecasting exercises such as Delphi, where having a diverse group of expertise can help to improve the quality and robustness of forecasts \citep{kupferschmidt}. These personas included: agronomist, economist, global affairs specialist, and the average Canadian. The specific expertise captured by these personas was selected to emulate the expertise of the larger CFPR research team.  Consensus between models was then reached by selecting any variable that scored 7 or higher. For the full prompt structure, refer to Appendix \ref{appendix:likert_prompting}.

\subsubsection{Additional Inputs for LLM Forecasters}
When using LLMs as direct forecasting models (i.e. LLMP and Direct Prompting), additional inputs were provided during prompting. These models' flexibility in handling diverse input formats allowed for the integration of supplementary information, including exogenous regressors, future forecasts, and historical CFPR data.

\textit{\textbf{1) Exogenous Regressors}}: Four human-selected time series that were thought to be major drivers were chosen for each of the target categories. 

\textit{\textbf{2) Future Forecasts}}: Three versions of future forecasts were generated and fed into the LLM as an additional input. These forecasts were produced by two local statistical models (Seasonal Naive and Exponential Smoothing) and one global model (TemporalFusionTransformer) with the same setup. The forecasts included forecasts generated across all evaluation windows as well as forecasts for the entire forecasting horizon of 18 months into the future. 

\textit{\textbf{3) Previous Releases of Canada's Food Price Report (CFPR)}}: Each year, a PDF document of the CFPR is made publicly available\footnote{\url{https://www.dal.ca/sites/agri-food/publications.html}}. For each forecast, each model was provided with a version of the content in the PDF. As GPT-4o was able to accept multimodal inputs, the report was incorporated directly as a PDF file whereas for the other commercial models (Gemini 1.5 Pro and Claude 3.5~Sonnet), the PDF files were converted using the PDF Reader Python library\footnote{\url{https://pypi.org/project/pdfreader/}}.

\subsection{Models}

Several models were evaluated for their ability to produce forecasts for the 9 food related CPI categories. These models were clustered into types to describe their overall modeling approach as well as their level of sophistication. The sophistication of models can refer to their number of parameters but can also increase based on the number and diversity of examples they experience during training. 

\textit{\textbf{Statistical Models}} -  These models extract patterns such as trend and seasonality directly from the data to produce future predictions. They rely on relatively simple mathematical formulations and are well-suited for datasets with clear and consistent temporal patterns. Examples include Seasonal Naive and Exponential Smoothing models \citep{Hyndman2018-fa}.

\textit{\textbf{Deep learning}} - These models use neural networks to capture complex patterns in the data, including nonlinear relationships and interactions between regressors. They typically require larger datasets and more computational resources compared to statistical models but can outperform simpler approaches when rich covariate information is available \citep{makridakis2018m4}. 

\textit{\textbf{Transformer}} - These models leverage the self-attention mechanism to model relationships across long sequences of data, making them effective for capturing both local and global dependencies in time-series data. Transformer-based models, such as the Temporal Fusion Transformer (TFT), are particularly well-suited for multivariate time-series forecasting where temporal and contextual regressors are known to play a significant role \citep{lim2021temporal}.

\textit{\textbf{Foundation}} - These models often leverage the transformer architecture but are pre-trained on extensive external datasets enabling them to adapt to a wide range of tasks. In particular, they perform well in "zero-shot" settings, where limited historical data is available \citep{bommasani2021opportunities, ansari2024chronos}. 

\textit{\textbf{LLM}} - LLMs are a form of foundation model that leverage extensive corpa of text-based data during pre-training. Recent research has suggested that these language-based models can be used to generate forecasts when prompted with descriptive and numerical inputs. Unlike conventional models, they can interpret and incorporate supplementary information, such as contextual descriptions, related regressors, and historical narratives (e.g., previous CFPR reports) \citep{requeima2024llm, williams2024context}.

\subsubsection{Time-series forecasting models} 
 For each CPI food category, different classes of models were implemented. All models, other than the LLM models were implemented through the AutoGluonTS library (v1.0)\footnote{\url{https://github.com/autogluon/autogluon}}. AutoGluon is an AutoML approach that allows for easy and direct comparison between multiple model types. The evaluated models ranged from simple statistical models such as \textit{ARIMA} or \textit{Exponential Smoothing} to high capacity ML models such as \textit{TemporoSpatialTransformer} \citep{lim2021temporal} and \textit{DeepAR} \citep{Salinas2020-sl}. Furthermore, Chronos, a pre-trained time-series foundation model was also included for zero-shot forecasting \citep{ansari2024chronos}. Each of the models was trained using several training paradigms including local (i.e.~only including the target variable), or globally (i.e.~a global multitask training objective over target variables and regressors) with or without covariates (Table \ref{tab:model_training_summary}). Furthermore, several dataset curation techniques including stratifying regressors by theme or through LLM curation were also evaluated (refer to Appendix \ref{appendix:groupings of variables}).

\subsubsection{LLMs as forecasting models}
Recent works have demonstrated the value of using LLMs for time-series forecasting \citep{jin2023large}. A particularly appealing advantage of these approaches is the relative ease and demonstrated success of providing additional contextual information at inference time {\citep{requeima2024llm, williams2024context}. Using the \textit{Direct Prompt} framework applied in ({\citet{ williams2024context}), three commercial LLMs were prompted to directly output a forecast as a structured output for some specified forecasting horizon (i.e.~18 months).  The prompt structure can be found in Appendix \ref{appendix:LLMP Prompts}. Despite the ability to generate stochastic forecasts to generate uncertainty estimates, all instances were run deterministically (hyperparameter temperature=0) for this initial set of experiments. 

 These models were used to forecast 3 of the categories of food pricing: (1) Bakery and Cereal, (2) Meat, and (3) Vegetables. Due to resource constraints, experimentation was limited to categories that have historically been challenging to predict and have experienced substantial fluctuation in recent years. For the LLM experiments, several strategies for incorporating context were tested including historical exogenous regressors, future forecasts generated by statistical and ML-based models, and the content of previous CFPRs.

\begin{table}[h!]
    \centering
    \scriptsize 
    \setlength{\tabcolsep}{2pt} 
    \caption{Summary of model types, training objectives, and data composition}
    \label{tab:model_training_summary}
    \resizebox{\linewidth}{!}
{
    \begin{tabular}
        {>{\centering\arraybackslash}m{2.2cm} 
         >{\centering\arraybackslash}m{2.7cm} 
         >{\centering\arraybackslash}m{1cm} 
         >{\centering\arraybackslash}m{1cm} 
         >{\centering\arraybackslash}m{1.2cm} 
         >{\centering\arraybackslash}m{1.3cm} 
         >{\centering\arraybackslash}m{1cm} 
         >{\centering\arraybackslash}m{1cm} 
         >{\centering\arraybackslash}m{1.3cm} 
         >{\centering\arraybackslash}m{1.3cm}}
        \toprule
        \multirow{6}{*}{\textbf{Model Family}} & \multirow{6}{*}{\textbf{Model Type}} & \multicolumn{8}{c}{\textbf{Training Objectives}} \\
        \cmidrule(lr){3-10}
         & & \textbf{Local} & \multicolumn{4}{c}{\textbf{Global}} & \multicolumn{3}{c}{\textbf{In-Context Learning}} \\
        \cmidrule(lr){3-3} \cmidrule(lr){4-7} \cmidrule(lr){8-10}
         & & \textbf{Local} & \textbf{CPI only} & \textbf{Theme only (x4)} & \textbf{LLM Selected only} & \textbf{All} & \textbf{Local} & \textbf{Forecasts} & \textbf{CFPR} \\
        \midrule
        \multirow{4}{*}{\textbf{Statistical}} & Naive & \cmark & \myxmark & \myxmark & \myxmark & \myxmark & \myxmark & \myxmark & \myxmark \\
        & Seasonal Naive & \cmark & \myxmark & \myxmark & \myxmark & \myxmark & \myxmark & \myxmark & \myxmark \\
        & AutoARIMA & \cmark & \myxmark & \myxmark & \myxmark & \myxmark & \myxmark & \myxmark & \myxmark \\
        & AutoETS & \cmark & \cmark & \cmark & \cmark & \cmark & \myxmark & \myxmark & \myxmark \\
        \midrule
        \multirow{3}{*}{\makecell{\textbf{Deep} \\ \textbf{Learning}}} & DeepAR & \cmark & \cmark & \cmark & \cmark & \cmark & \myxmark & \myxmark & \myxmark \\
        & SimpleFeedForward & \cmark & \cmark & \cmark & \cmark & \cmark & \myxmark & \myxmark & \myxmark \\
        & DLinear & \cmark & \cmark & \cmark & \cmark & \cmark & \myxmark & \myxmark & \myxmark \\
        \midrule
        \multirow{2}{*}{\textbf{Transformer}} & Temporal Fusion Transformer & \cmark & \cmark & \cmark & \cmark & \cmark & \myxmark & \myxmark & \myxmark \\
        & PatchTST & \cmark & \cmark & \cmark & \cmark & \cmark & \myxmark & \myxmark & \myxmark \\
        \midrule
        \textbf{Foundation} & Chronos & \cmark & \cmark & \cmark & \cmark & \cmark & \myxmark & \myxmark & \myxmark \\
        \midrule
        \multirow{3}{*}{\textbf{LLM}} & Claude Sonnet 3.5 & \myxmark & \myxmark & \myxmark & \myxmark & \cmark & \cmark & \cmark & \cmark \\
        & Gemini 1.5 Pro & \myxmark & \myxmark & \myxmark & \myxmark & \cmark & \cmark & \cmark & \cmark \\
        & GPT-4o & \myxmark & \myxmark & \myxmark & \myxmark & \cmark & \cmark & \cmark & \cmark \\
        \bottomrule
    \end{tabular}
    }
\end{table}

\subsection{Model Evaluation and Final Forecast Generation}
Historical monthly CPI data for each food category from 1986 to 2016 were used to train the individual models. Given the annual periodic nature of the CFPR, the models were evaluated based on their ability to produce 18-month forecasts for the six most recent years (2018 to 2023). For all non-LLM models, the context length remained consistent at 36 months. Models were evaluated for their ability to complete a univariate forecasting task of predicting a single food CPI category at a time. Performance metrics were calculated using MAPE scores between the forecast values and ground truths across the entire test period. 

Upon completion of experiments, final forecasts were generated through a variable size (1-3) ensemble for each category. This was completed by exploring all combinations of the top 10 performing models for each category and selecting the top-performing ensemble during the same evaluation period.

\section{Results}
We evaluated how different types of context affected forecasting performance, focusing on models capable of incorporating exogenous time series (Figure \ref{fig:contextcompare}). For baseline comparisons, we experimented with two settings: models trained with all available exogenous regressors and models trained only on the target time series (no exogenous regressors). 

\begin{figure}
    \centering
    \includegraphics[width=1\linewidth]{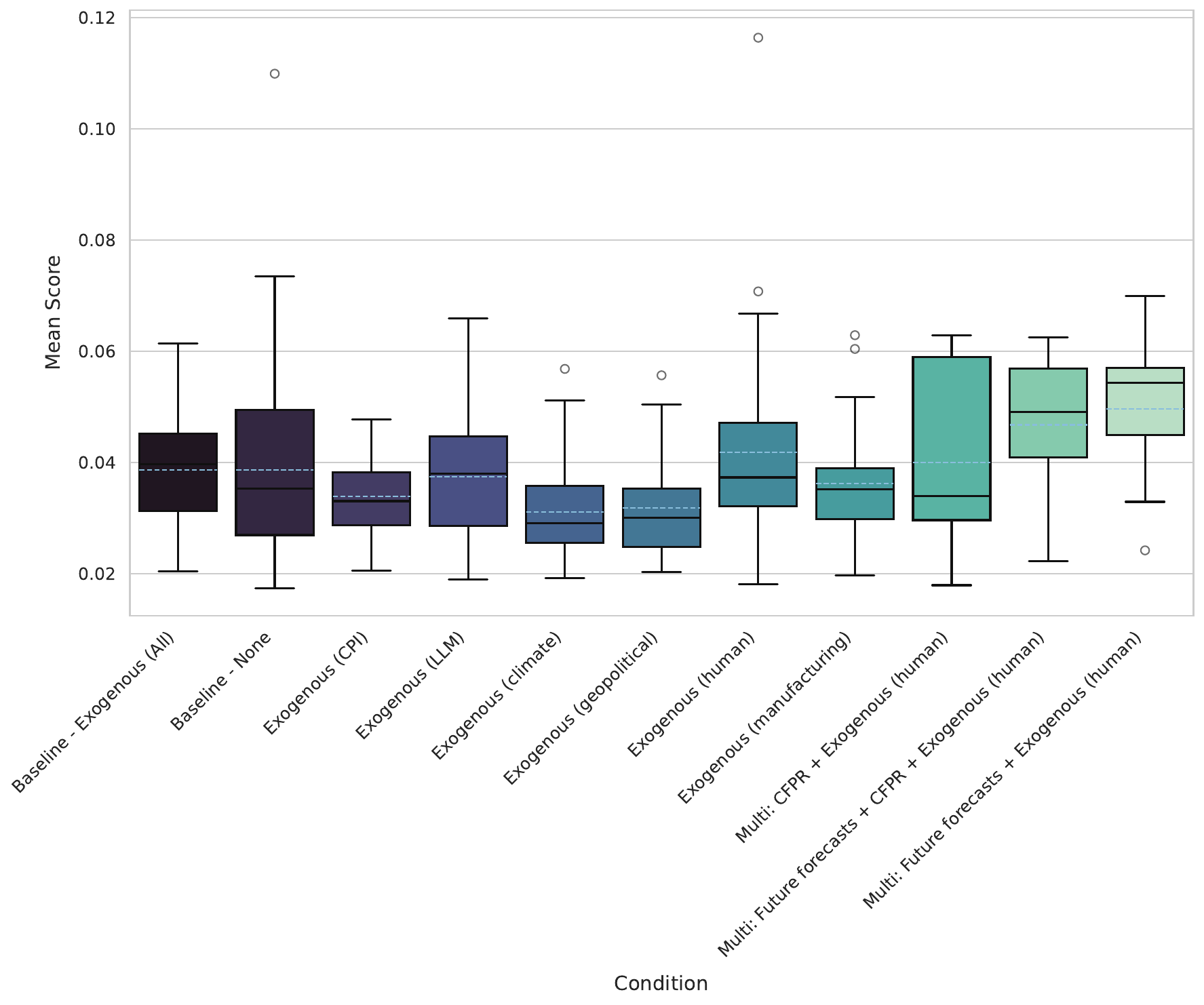}
    \caption{Performance changes for different context inclusion techniques -- Averaged over all food categories and model types}
    \label{fig:contextcompare}
\end{figure}

When averaged across all food categories and model types, only marginal improvements can be observed through data curation. While including all exogenous regressors tended to deteriorate performance, it appeared that overall, even in cases where exogenous regressors were curated, there was not a significant improvement over a purely local forecasting approach.  However, it does appear that model consistency improved when leveraging curated approaches. Overall, the best-performing selection techniques used exclusively climate regressors and geopolitical regressors which is in line with general discussion within the food economics community \citep{Kalkuhl2016-xc}. 

When considering the LLM forecasters, it appeared that the inclusion of future forecasts actually deteriorated performance. This could be because of documented forecasting inaccuracy during the evaluation periods of 2021-2022 where food prices were extremely volatile \citep{Kupferschmidt2024-lx}. In general, the condition where the previous edition of the CFPR was provided (i.e.~Multi: CFPR + Exogenous (human)) resulted in MAPE scores comparable to those selected by themes. However, the incorporation of these multimodal inputs resulted in a notable decrease in stability observed through the high variance between evaluation periods and food categories.

To better understand these findings, the same comparison was broken down by food category (Table \ref{tab:performance_metrics_inclusion}). The results suggest that there was no food category that benefited from including all of the regressors during global training. Furthermore, all of the best-performing conditions included some curated set of covariates and outperformed baselines where no exogenous regressors were provided. Consistent with averaged results, the most commonly top-performing groupings of regressors represented climate, geopolitical, or food CPI time series. When looking exclusively at the categories that included multi-modal inputs, there appeared to be mixed results for the inclusion of the CFPR. For Bakery and Meat, the addition of the previous year's CFPR tended to improve performance, however, for vegetables the best food CPI regressors provided the most useful context.

\begin{table}[!t]
\caption{MAPE for various methods of providing context organized by target task. Bold represents the best context inclusion technique for that category averaged across different evaluation windows and model types.}
\label{tab:performance_metrics_inclusion}
\centering
\resizebox{\textwidth}{!}{  

\footnotesize  
\begin{tabular}{lccccccccccc}
\toprule
\multirow{2}{*}{\textbf{Category}} & \multicolumn{2}{c}{\textbf{Baseline}} & \multicolumn{6}{c}{\textbf{Exogenous}} & \multicolumn{3}{c}{\textbf{Multi (including Exogenous)}} \\
\cmidrule(lr){2-3} \cmidrule(lr){4-9} \cmidrule(lr){10-12}
 & \textbf{All} & \textbf{None} & \textbf{CPI} & \textbf{LLM} & \textbf{Climate} & \textbf{Geopolitical} & \textbf{Manufact.} & \textbf{Human} & \textbf{CFPR} & \textbf{Future + CFPR} & \textbf{Future} \\
\midrule
Bakery & 0.045 ± 0.01 & 0.044 ± 0.01 & 0.043 ± 0 & 0.045 ± 0.01 & 0.039 ± 0.00 & 0.038 ± 0.00 & 0.042 ± 0.01 & 0.040 ± 0.01 & \textbf{0.034 ± 0.00} & 0.044 ± 0.01 & 0.045 ± 0.01 \\
Dairy & 0.034 ± 0.01 & 0.032 ± 0.01 & 0.029 ± 0 & 0.033 ± 0.01 & \textbf{0.026 ± 0.00} &\textbf{ 0.026 ± 0.01} & 0.031 ± 0.00 & 0.034 ± 0.01 & - & - & - \\
Fish & 0.030 ± 0.01 & 0.025 ± 0.01 & \textbf{0.021 ± 0} & 0.028 ± 0.01 & 0.023 ± 0.01 & 0.024 ± 0.00 & 0.025 ± 0.01 & 0.031 ± 0.01 & - & - & - \\
Food & 0.031 ± 0.01 & 0.034 ± 0.01 & 0.028 ± 0 & 0.031 ± 0.01 & 0\textbf{.023 ± 0.00} & 0.025 ± 0.00 & 0.028 ± 0.01 & 0.032 ± 0.01 &  & - &  -\\
Fruit & 0.041 ± 0.00 & 0.041 ± 0.01 & \textbf{0.030 ± 0 }& 0.036 ± 0.01 &\textbf{ 0.029 ± 0.00} & 0.030 ± 0.00 & 0.036 ± 0.00 & 0.038 ± 0.01 &-  &-  & - \\
Meat & 0.035 ± 0.01 & 0.031 ± 0.01 & 0.037 ± 0 & 0.035 ± 0.01 & 0.028 ± 0.01 & 0.029 ± 0.01 & 0.033 ± 0.01 & 0.032 ± 0.01 & \textbf{0.025 ± 0.01} & 0.039 ± 0.01 & 0.042 ± 0.02 \\
Other & 0.040 ± 0.01 & 0.043 ± 0.01 & 0.036 ± 0 & 0.040 ± 0.01 & \textbf{0.032 ± 0.00} & 0.032 ± 0.01 & 0.037 ± 0.00 & 0.040 ± 0.01 & - & - & - \\
Vegetables & 0.054 ± 0.01 & 0.060 ± 0.02 &\textbf{ 0.048 ± 0} & 0.052 ± 0.01 & 0.050 ± 0.01 & 0.051 ± 0.01 & 0.058 ± 0.01 & 0.070 ± 0.02 & 0.061 ± 0.00 & 0.058 ± 0.01 & 0.062 ± 0.01 \\
\bottomrule
\end{tabular}
}
\end{table}

\subsection{Do different food categories favor different families of models?}
Both local and global models were evaluated for their ability to produce forecasts over each of the evaluation periods. Figure \ref{fig:model_class_ordering} shows the best performing configuration (Rank=1) for each model family in order of increasing complexity (left to right). Each model was classified into one of four model classes: statistical, deep-learning, transformer, and foundation models.  

\begin{figure}
        \centering
        \includegraphics[width=0.8\linewidth]{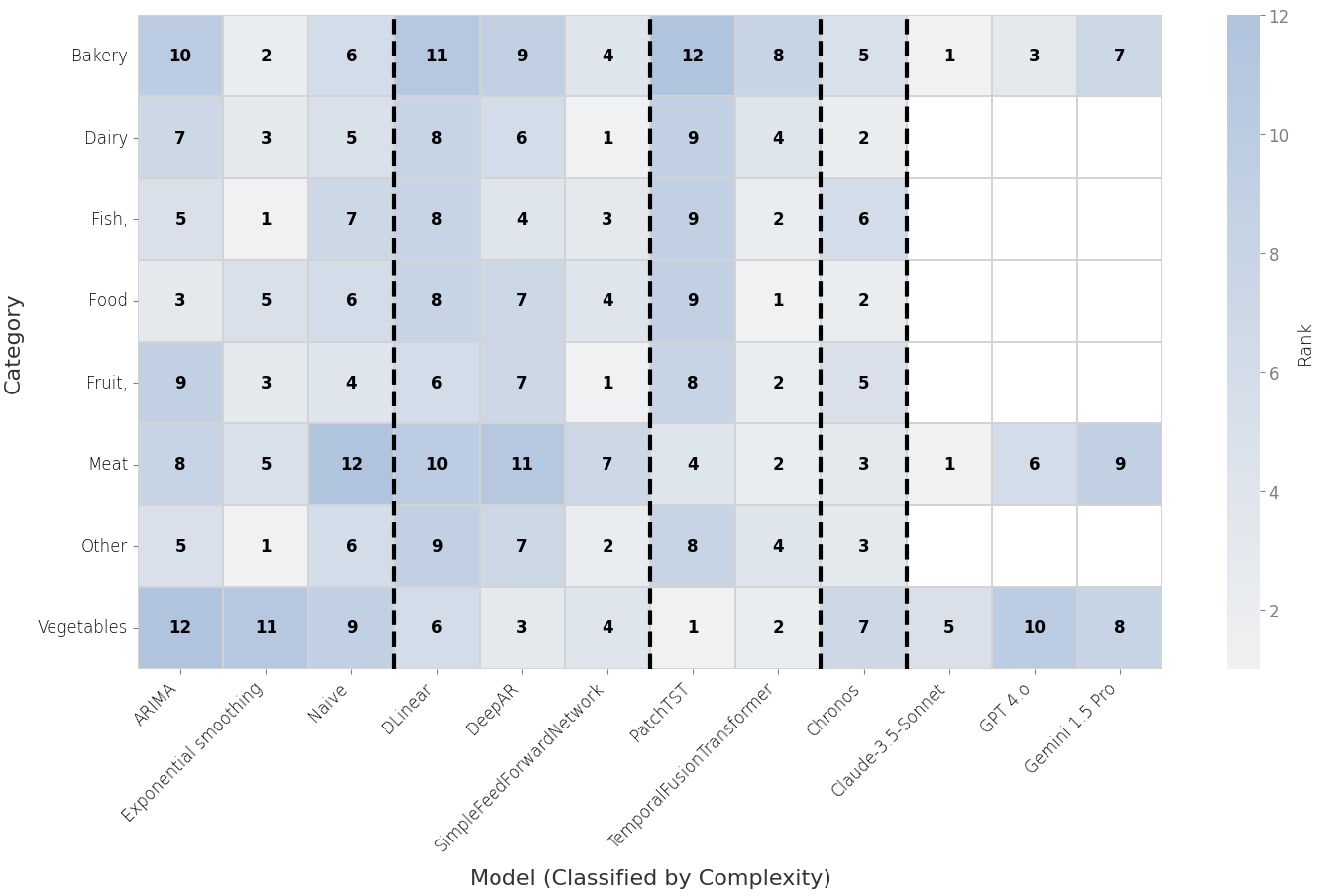}
        \caption[Model rankings for each food category in CPI sorted in order of increasing complexity of model classes. Up-facing arrows indicate that higher numbers represent more complexity for a given category, while down-facing arrows indicate that lower numbers represent more complexity. Results represent single best-performing regressors from each model]{Model rankings for each food category in CPI sorted in order of increasing complexity of model classes. Results represent single best-performing regressors from each model. Models are sorted in order of capacity increasing from left to right with model classes being statistical, deep-learning-based, transformer-based, foundation models, and LLMs. Model capacity was estimated based on best estimates of \# of parameters. }    \label{fig:model_class_ordering}
\end{figure}

To better understand what characteristics could determine if a specific forecasting target would be better suited for a different model class, the ranked complexity of each CPI category was calculated. Five complexity metrics including trend and seasonality strength, residual variance, mean average deviation (MAD) and Shannon Entropy were used to quantify different aspects of complexity. Values for each metric were calculated using 3-year overlapping windows from 1986-2024. We computed the average relative ranking for each of the evaluated metrics, and ranked the categories based on this average value. The residual variance represents the variance of residual components left after the seasonality and trend are extracted from the signal and are used as a representation of random fluctuations present in the signal. Similarly, the residual mean absolute deviation (MAD) calculates the deviation of the residuals from zero providing a measure of the residual's magnitude. 

\begin{table}[!t]
\caption{Metrics for various target task complexity, including trend strength, seasonality strength, residual variance, residual MAD, Shannon's entropy, and overall rank. Arrows indicate the direction of increased complexity.}
\label{tab:metrics_food_categories}
\centering
\resizebox{\textwidth}{!}{  
\footnotesize  
\begin{tabular}{lcccccc}
\toprule
\textbf{Category} & \textbf{Trend Strength $\downarrow$} & \textbf{Seasonality Strength$\downarrow$ } & \textbf{Residual Variance $\uparrow$} & \textbf{Residual MAD $\uparrow$} & \textbf{Shannon's Entropy $\uparrow$} & \textbf{Overall Rank $\uparrow$} \\
\midrule
Bakery & 0.952 ± 0.064 & 0.798 ± 0.116 & 0.263 ± 0.364 & 0.33 ± 0.241 & 4.707 ± 0.207 & 4 \\
Dairy & 0.944 ± 0.115 & 0.813 ± 0.112 & 0.11 ± 0.168 & 0.21 ± 0.13 & 4.559 ± 0.33 & 1 \\
Fish & 0.917 ± 0.123 & 0.797 ± 0.12 & 0.172 ± 0.134 & 0.296 ± 0.121 & 4.616 ± 0.337 & 5 \\
Restaurants & 0.998 ± 0.002 & \textbf{0.721 ± 0.171} & 0.022 ± 0.036 & 0.094 ± 0.063 & 4.896 ± 0.198 & 3 \\
Fruit & 0.843 ± 0.141 & 0.866 ± 0.103 & 0.862 ± 0.666 & 0.677 ± 0.26 & 4.845 ± 0.167 & 6 \\
Meat & 0.924 ± 0.113 & 0.76 ± 0.139 & 0.394 ± 0.512 & 0.425 ± 0.263 & 4.71 ± 0.226 & 7 \\
Other food & 0.95 ± 0.067 & 0.859 ± 0.08 & 0.13 ± 0.129 & 0.253 ± 0.128 & 4.565 ± 0.293 & 2 \\
Vegetables & \textbf{0.736 ± 0.245} & 0.908 ± 0.066 & \textbf{3.698 ± 3.685} &\textbf{ 1.341 ± 0.609} &\textbf{ 5.019 ± 0.088} & \textbf{8} \\
\bottomrule
\end{tabular}
}
\end{table}

In general, the more complex time CPI categories, such as Vegetables and Meat, tended to select higher-capacity models, like transformer-based models. In contrast, simple categories, such as Dairy, Fish and Other, which had extremely low Residual Variance and MAD, performed best with simple Exponential Smoothing models or simpler Deep Learning models such as SimpleFeedForwardNetwork. This coincides with the fact that categories with high Residual variance and MAD such as Vegetables and Fruit performed poorly with all statistical models, wich depend heavily on signal stationarity.

 Consistent with findings from similar approaches, such as those reported by \citet{Kristina-L-Kupferschmidt-Cody-Kupferschmidt-Joshua-A-Skorburg-Branka-Agic-Tara-Elton-Marshall-Hayley-Hamilton-Gina-Stoduto-Katherine-Vink-Samantha-Wells-Christine-Wickens-Graham-W-TaylorUnknown-ra}, the results show that model performance—and their relative rankings—are highly dependent on the context window provided at inference time. This context captures both intrinsic properties of specific categories (e.g., interactions between targets and regressors) and behaviors influenced by temporal perturbations (e.g., recessions, pandemics). Notably, no single model consistently outperformed simple statistical approaches, such as exponential smoothing, across all time periods. This suggests that while certain contexts may benefit from highly complex models, others may be better suited to simpler models that rely primarily on historical target time-series.
 
 The idea of encoding the initial context provided to predict the best-performing model is reminiscent of Mixture-of-Expert (MoE) models in NLP, where inputs are routed to specialized expert models based on intrinsic features of the context. Future research could investigate parallels between language and time-series data, exploring how intrinsic characteristics could serve as routing signals to the best expert model. Taking inspiration from language sequences (e.g.~syntax, semantics), time-series data characteristics (e.g.~trend, seasonality, dimensionality) could be used by routers to extract meaningful embeddings from inputs. Such an approach could improve model specificity, particularly in rapidly changing contexts where both traditional and ML time-series models often struggle to provide consistent accuracy.

\subsection{What models were selected for the 2025 food price report?}
Probabilistic forecasts were generated for each of the 9 CPI categories (Figure \ref{fig:CFPR2025_forecasts}). The forecasts were an ensemble of size 1-3 (determined combinatorially) based on the best performance during the evaluation periods. 

\begin{figure}
    \centering
    \includegraphics[width=1\linewidth]{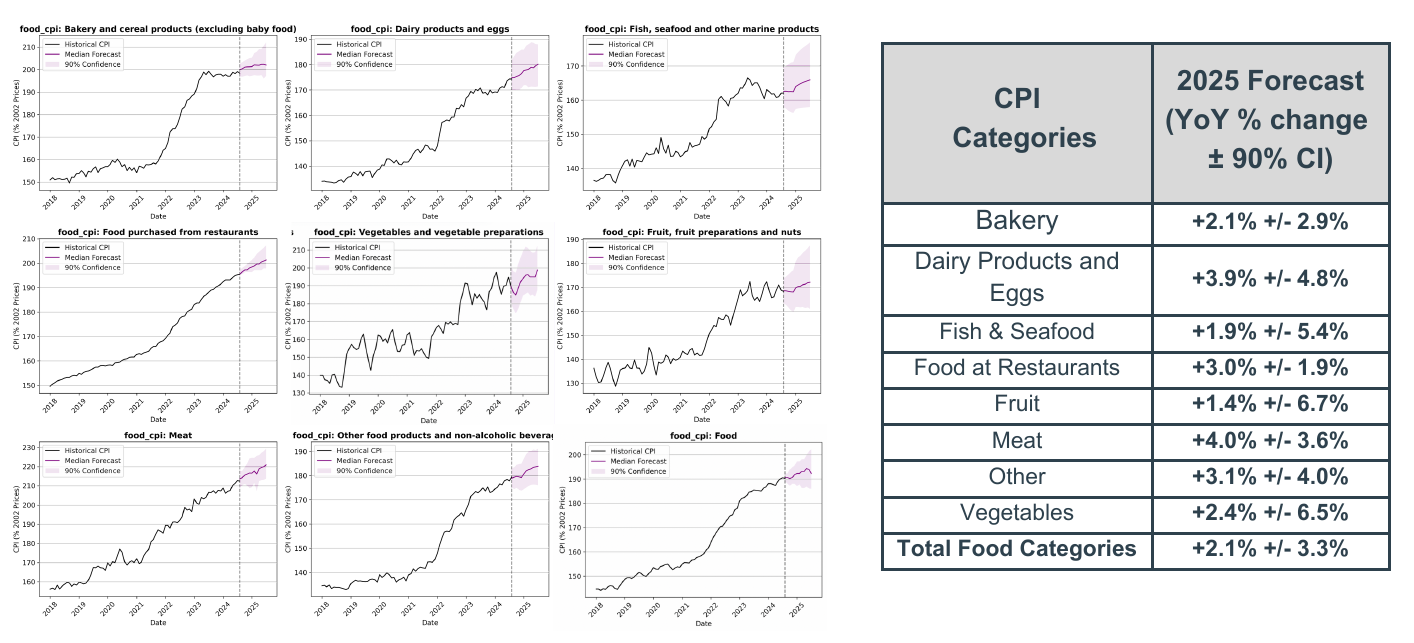}
    \caption{Forecasts and projected year over year (2025/2024) \% change for the food CPI categories included in Canada's Food Price Report}
    \label{fig:CFPR2025_forecasts}
\end{figure}

An analysis was completed to understand the proportion of top-scoring models that were included in the ensemble fell into the different model categories (Table \ref{tab:modelclas_freq_forecasts}). Only two food categories (Other and Fish) utilized statistical models in this year's report, whereas the vast majority of models included were ML-based. This differs from previous years where statistical models often outperformed deep learning approaches. All categories used an ensemble with the exception of Food and Restaurants, both of which performed best using global variants of Chronos, a single time-series foundation model. 

\begin{table}[h!]
\caption{Inclusion of model classes in final ensembles for 2025 Forecasts}
\label{tab:modelclas_freq_forecasts}
\footnotesize
\centering
\begin{tabular}{lccccc}
\toprule
\textbf{Category}    & \textbf{Statistical model} & \textbf{Deep Learning} & \textbf{Transformer model} & \textbf{Foundation model} & \textbf{LLM Forecaster} \\ \midrule
Bakery       & \myxmark       & \myxmark       & \cmark      & \cmark      & \cmark        \\ 
Dairy        & \myxmark       & \myxmark       & \cmark      & \cmark      & -      \\ 
Fish         & \cmark         & \myxmark       & \cmark      & \myxmark    & -      \\ 
Restaurants  & \myxmark       & \myxmark       & \myxmark    & \cmark      & -     \\ 
Food         & \myxmark       & \myxmark       & \myxmark    & \cmark      & -      \\ 
Fruit        & \myxmark       & \cmark         & \cmark      & \myxmark    & -      \\ 
Meat         & \myxmark       & \myxmark       & \cmark      & \myxmark    & \cmark        \\ 
Other        & \cmark         & \cmark         & \myxmark    & \cmark      & -      \\ 
Vegetables   & \myxmark       & \myxmark       & \cmark      & \cmark      & \myxmark      \\ \bottomrule
\end{tabular}
\end{table}

Similarly, results were analyzed to understand which context inclusion strategies appeared most prominently in the final ensembled forecasts (Table \ref{tab:data_freq_forecasts}). Several categories (Fruit, Meat, and Other) appeared to benefit from the use of a geopolitical-specific model. Conversely, only Fruit benefited from a climate-exclusive model despite there being a known link between major climate events and the pricing of other categories such as Bakery and Vegetables. Importantly, this does not imply that climate regressors are unimportant as exogenous regressors for these categories. In fact, climate regressors were included in certain other model configurations, such as the LLM-suggested exclusive model for Vegetables. Additionally, several categories included at least one globally trained model that incorporated all potential regressors, representing climate, geopolitical, and economic factors comprehensively.

\begin{table}[h!]
\caption{Inclusion of data curation methods in final ensembles for 2025 Forecasts}
\label{tab:data_freq_forecasts}
\footnotesize
\centering
\renewcommand{\arraystretch}{1.2} 
\begin{tabular}{m{1.6cm}m{1.6cm}m{1.7cm}m{1.6cm}m{1.3cm}m{1.8cm}m{2.4cm}m{1.2cm}}
\hline
\textbf{Category}    & \textbf{All \newline Regressors} & \textbf{No \newline Regressors} & \textbf{CPI Food \newline Categories} & \textbf{Climate Only} & \textbf{Geopolitical Only} & \textbf{LLM Suggested \newline Only} & \textbf{2024 CFPR} \\ \hline
Bakery       & \checkmark & \checkmark & \myxmark       & \myxmark       & \myxmark       & \myxmark       & \checkmark         \\ 
Dairy        & \checkmark & \checkmark & \myxmark       & \myxmark       & \myxmark       & \myxmark       & -           \\ 
Fish         & \myxmark   & \checkmark & \checkmark     & \myxmark       & \myxmark       & \checkmark     & -           \\ 
Restaurants  & \checkmark & \myxmark   & \myxmark       & \myxmark       & \myxmark       & \myxmark       & -         \\ 
Food         & \checkmark & \myxmark   & \myxmark       & \myxmark       & \myxmark       & \myxmark       & -          \\ 
Fruit        & \myxmark   & \myxmark   & \myxmark       & \checkmark     & \checkmark     & \checkmark     & -           \\ 
Meat         & \myxmark   & \checkmark & \myxmark       & \myxmark       & \checkmark     & \myxmark       & \checkmark         \\ 
Other        & \checkmark & \checkmark & \myxmark       & \myxmark       & \checkmark     & \myxmark       & -           \\ 
Vegetables   & \myxmark   & \checkmark & \myxmark       & \myxmark       & \myxmark       & \checkmark     & \myxmark           \\ \hline
\end{tabular}
\end{table}

Based on these results, it appears that several of the methodological changes incorporated in the 2025 CFPR, including the incorporation of two new model classes: foundation models and LLMs, and inclusion of new data curation strategies outperformed other modeling strategies. Future work should continue to explore prompting strategies for best-leveraging LLMs to better understand the forecasting problem at hand and continue to advise on data that may be useful. 

\section{Conclusions}
This document serves as an accompanying technical report for the 2025 CFPR, detailing the experimental methodology behind our forecasts and exploring data-centric approaches to model development. The findings suggest that simple curation techniques, such as thematic groupings of exogenous regressors, can improve model performance.  Future research should further explore additional prompt strategies for leveraging LLMs as data-curation tools, particularly examining how model performance changes with increased contextual information.

We also hypothesize that some intrinsic properties of time series may influence model class selection. This highlights the potential to adapt modern language mixture-of-expert (MoE) models to the time-series domain. By combining data-centric techniques such as data-curation with new methods of incorporating context into forecasting decisions, we found that forecasting error can be reduced, often outperforming statistical and fully globally trained models (i.e., modeling strategies that include all possible regressors). 

This work builds upon previous forecasting efforts to anticipate changes in Canada's food prices, an issue that become particularly pertinent given the recent economic impacts of food inflation in Canada.

\bibliographystyle{plainnat}
\bibliography{references.bib}

\newpage
\input{Appendix/Appendix}

\end{document}

%% file: Appendix/Appendix.tex
\appendix
\section*{Appendix}
\label{appendix:Chapter8}
\addcontentsline{toc}{section}{Appendix}

\section{LLM Likert Prompting using Personas}
\label{appendix:likert_prompting}

\subsubsection*{Guardrails to Promote Clean Slate Prompting from Prompt to Prompt}
\begin{itemize}
    \item Memory turned off
    \item ``Improve the model for everyone'' turned off
    \item ``Delete all chats'' after each prompt
    \item Clear memory after each prompt
\end{itemize}

\subsubsection*{Personas}
\begin{itemize}
    \item \textbf{Food Economist}: You are an experienced Food Economist with a deep understanding of the Canadian food economy. You have been hired to consult on Canada’s Food Price Report because of your knowledge and expertise in Canada’s food market.
    
    \item \textbf{Agronomist}: You are an experienced Agronomist specializing in North American agriculture. You have been hired to consult on Canada’s Food Price Report because of your knowledge and expertise in agriculture and its impacts on Canadian food.
    
    \item \textbf{Global Affairs Specialist}: You are an experienced Global Affairs Specialist with a deep understanding of Canada’s global relations. You have been hired to consult on Canada’s Food Price Report because of your knowledge and expertise in Canadian-Global contexts.
    
    \item \textbf{Average Canadian}: You are an average Canadian citizen who does their own grocery shopping, budgeting, and works an average Canadian job. You are being surveyed for Canada’s Food Price Report because of your experience living, working, and grocery shopping in Canada.
\end{itemize}

\subsubsection*{General Prompt}
\noindent For Canada's Food Price Report 2025, we are looking to better understand the factors that may contribute to fluctuations in Canadian food prices.

\noindent We have collected 30 monthly time-series variables and have described them below. We are asking you to rank them in terms of their influence on food prices in Canada according to a scale from 1 -- ``Not at all influential'' to 9 -- ``Extremely influential''.

\noindent Please rank each variable in terms of its influence on food prices in Canada according to the following scale:

\begin{itemize}
    \item 1 -- Not at all influential
    \item 2
    \item 3 -- Slightly influential
    \item 4
    \item 5 -- Moderately influential
    \item 6
    \item 7 -- Very influential
    \item 8
    \item 9 -- Extremely influential
\end{itemize}

\noindent \textbf{\{Variable list\}}

\subsection*{Prompt for Different Personas}
\noindent \textbf{\{Insert persona description here\}}

\noindent For Canada's Food Price Report 2025, we are looking to better understand the factors that may contribute to fluctuations in Canadian food prices.

\noindent We have collected 30 monthly time-series variables and have described them below. We are asking you to rank them in terms of their influence on food prices in Canada according to a scale from 1 -- ``Not at all influential'' to 9 -- ``Extremely influential''.

\noindent From the context of your role as \{role name\}, please rank each variable in terms of its influence on food prices in Canada according to the following scale:

\begin{itemize}
    \item 1 -- Not at all influential
    \item 2
    \item 3 -- Slightly influential
    \item 4
    \item 5 -- Moderately influential
    \item 6
    \item 7 -- Very influential
    \item 8
    \item 9 -- Extremely influential
\end{itemize}

\noindent \textbf{\{Variable list\}}

\subsection{Variable List}
Variables: 

Economic Policy Uncertainty Index:The Economic Policy Uncertainty Index tracks the level of uncertainty in economic policies in Canada by analyzing the frequency and content of newspaper articles discussing policy uncertainty. This index reflects how uncertain people are about future economic policies and their potential impact. 

Petroleum oil and grease production price index (U.S.):This index measures the price changes over time for petroleum lubricating oil and grease manufacturing in the U.S. It reflects the costs that producers face when manufacturing these products, and it helps track inflation within the petroleum industry. 

Food manufacturing price index (U.S.):The Food Manufacturing Price Index monitors the price changes for goods produced by the food manufacturing industry in the U.S. It provides insights into inflationary trends within the food sector by measuring how much manufacturers receive for their products over time. 

El Niño-Southern Oscillation (ENSO):The ENSO is a recurring climate phenomenon that involves periodic changes in sea surface temperatures in the central and eastern tropical Pacific Ocean. These temperature fluctuations significantly impact global weather patterns and climate conditions, leading to various environmental effects such as altered rainfall and temperature variations. 

Number of work stoppages (All industries):This variable counts the total number of work stoppages across all industries in Canada. Work stoppages refer to situations where groups of employees collectively cease work, typically due to disputes or negotiations with employers over labor conditions or contracts. 

U.S. imports from Mexico:This variable tracks the total value (in millions of dollars) of goods imported by the United States from Mexico. It provides a measure of trade flow between the two countries, reflecting the economic relationship and dependency on Mexican goods. 

U.S. Recession Probabilities:The smoothed U.S. recession probability tracks the chance (in percentages) that the United States is in a recession at any given time. This is calculated using a mathematical model (Markov-switching) and various U.S. economic variables (employment, industry production, income, and trade sales). 

Palmer Drought Severity Index (PDSI):The PDSI measures relative soil moisture conditions in primary American growing regions. It uses a standardized index based on a simplified soil water balance, helping to estimate the severity of drought conditions and their potential impact on agriculture and water resources. 

USD-CAD exchange rate:This variable tracks the exchange rate between the U.S. dollar and the Canadian dollar, indicating how many Canadian dollars are needed to purchase one U.S. dollar. It reflects the relative value of the two currencies and impacts trade and investment between the countries. 

Energy CPI:The Energy Consumer Price Index (CPI) for Canada measures the average change in prices paid by consumers for energy-related products and services. This index includes prices for items such as electricity, gasoline, and natural gas, providing insights into energy cost trends for Canadian households. 

Snow Water Equivalent (SWE):Snow Water Equivalent (SWE) measures the amount of water contained within the snowpack in California's growing regions. This metric is crucial for understanding water availability for agriculture, as it indicates the potential water supply from melting snow. 

Farm product price index:The Canadian Farm Product Price Index (FPPI) measures the changes in prices that farmers receive for their agricultural commodities, including crops, livestock, and animal products. It provides insights into market conditions and economic trends affecting Canadian agriculture. 

Federal Funds Effective Rate (U.S.):The Federal Funds Effective Rate (

Average price of electricity in U.S. cities:This variable measures the average price of electricity per kilowatt-hour in various U.S. cities, expressed in U.S. dollars. It provides an indication of electricity cost trends for urban consumers across the country. 

Interest rates:This variable measures the long-term government bond yields (10-year) for Canada. It reflects the cost of borrowing for the Canadian government and serves as a benchmark for other interest rates in the economy, influencing investment and economic growth. 

U.S. imports from China:This variable tracks the total value (in millions of dollars) of goods imported by the United States from China. It provides a measure of trade flow between the two countries, reflecting the economic relationship and dependency on Chinese goods. 

Pesticide and fertilizer production price index (U.S.):This index measures the price changes for products in the U.S. pesticide, fertilizer, and other agricultural chemical manufacturing industry. It tracks inflationary trends and cost variations within this critical agricultural sector. 

Crude Oil Price:The spot price of West Texas Intermediate (WTI) crude oil is measured in dollars per barrel. This variable reflects the market price for crude oil at a specific point in time, influencing energy costs and economic conditions globally. 

U.S. Unemployment Rate:The U.S. Unemployment Rate (\%) represents the percentage of the labor force that is unemployed. It is a key indicator of labor market health and overall economic conditions. 

Oil and gas machinery production price index (U.S.):This index measures the price changes for machinery and equipment used in U.S. oil and gas fields. It reflects cost trends and inflationary pressures within the oil and gas industry. 

U.S. imports from Canada:This variable tracks the total value (in millions of dollars) of goods imported by the United States from Canada. It provides a measure of trade flow between the two countries, reflecting the economic relationship and dependency on Canadian goods. 

Raw materials price index:The Raw Materials Price Index (RMPI) measures the price changes for raw materials purchased for further processing by manufacturers in Canada, excluding crude energy products. It indicates cost trends and inflationary pressures in the production process. 

Total U.S. employees (Non-farm):This variable measures the total number of workers in the U.S. economy, excluding farm employees, proprietors, private household employees, unpaid volunteers, and the unincorporated self-employed. It is a key indicator of labor market health and economic activity. 

Canola oil production:This variable measures the amount of canola oil produced in Canada, expressed in tonnes. It provides insights into the production levels and economic significance of canola oil in the Canadian agricultural sector. 

Fertilizer commodity price index:The world commodity price index for fertilizer tracks the price changes for fertilizer products globally, expressed in U.S. dollars. It reflects market conditions, supply and demand dynamics, and cost trends in the fertilizer industry. 

All Items CPI (U.S.):The Consumer Price Index (All Items) in U.S. city average measures the average change in prices paid by urban consumers for a broad range of goods and services. It is a key indicator of inflation and cost of living. 

Line haul railroads production price index (U.S.):This index measures the price changes for services provided by line-haul railroads in the U.S. It reflects cost trends and inflationary pressures within the freight transportation industry. 

Milk sold off farms:This variable measures the total volume of milk sold off Canadian farms for both fluid and industrial purposes, expressed in kilolitres. It indicates production levels and economic activity within the dairy sector. 

Nitrogenous fertilizer production price index (U.S.):This index measures the price changes for nitrogenous fertilizers manufactured in the U.S. It tracks cost trends and inflationary pressures within the agricultural chemical industry. 

Food commodity price index:The world commodity price index for food tracks the price changes for various food items, including oils, meals, grains, and others, expressed in U.S. dollars. It reflects global market conditions and cost trends in the food sector. 

\section{Grouping of data sources used for experiments}
\label{appendix:groupings of variables}

\begin{figure}[h]
    \centering
    \includegraphics[width=1\linewidth]{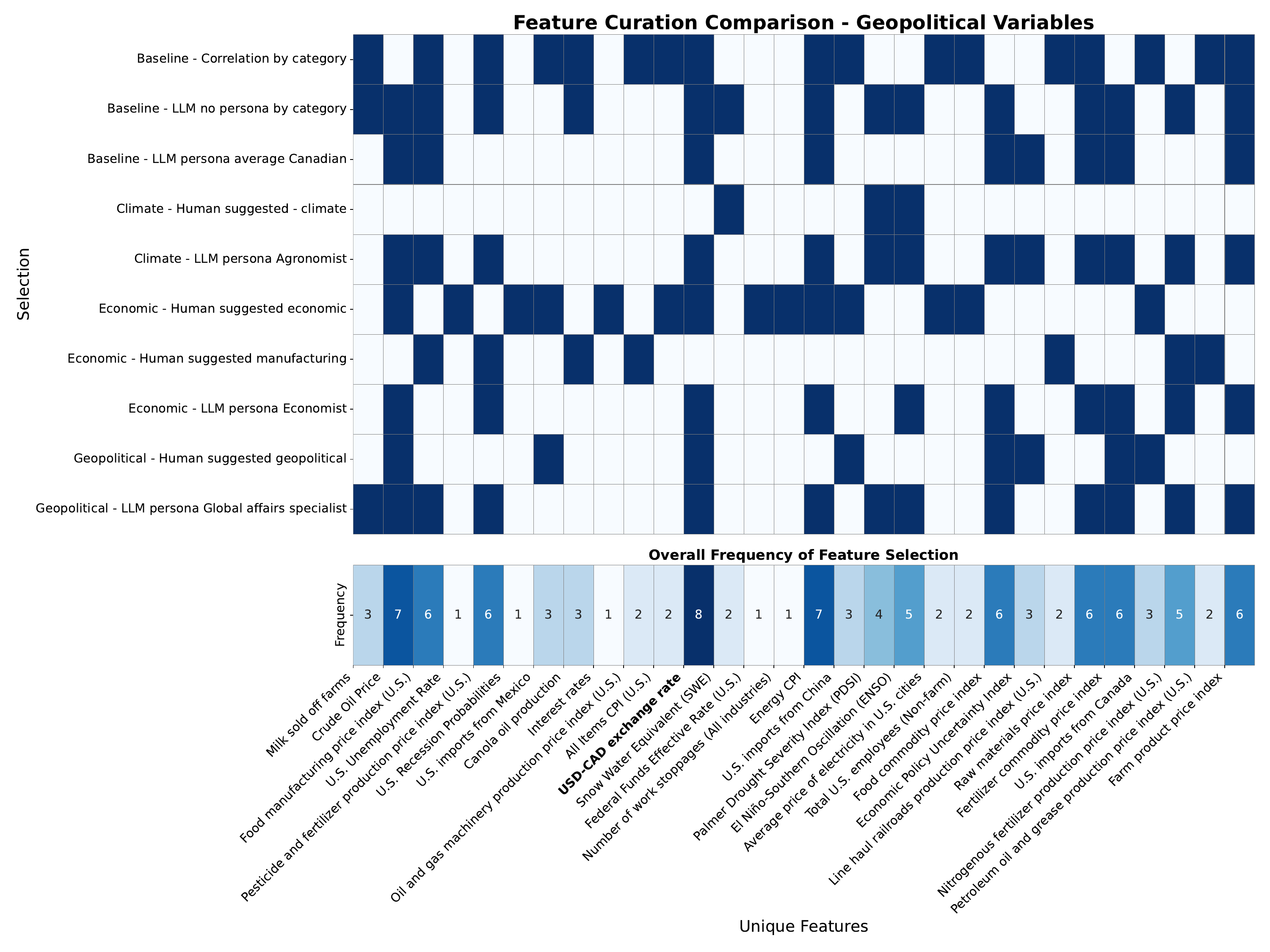}
    \caption{Inclusion of specific time-series in evaluated groupings}
    \label{fig:data_groupings}
\end{figure}

\section{LLM Prompting of Closed source models}
\label{appendix:LLMP Prompts}

\noindent \textbf{Task Description:}

I have a time series forecasting task for you. Please forecast the \texttt{target\_meat} variable described below. Please do not use Python and just directly predict the output.

\subsection*{Variable Descriptions:}
\begin{itemize}
    \item \texttt{target\_meat}: Food CPI measures changes in food prices experienced by Canadian consumers by comparing the cost of a fixed basket of food items (meat) over time.
    \item \texttt{exogenous\_fmpi}: Food manufacturing price index (U.S.) monitors price changes in U.S. food manufacturing reflecting sector inflation trends.
    \item \texttt{exogenous\_commodity\_food}: Food commodity price index tracks global price changes for various food items reflecting market conditions.
    \item \texttt{exogenous\_excaus}: USD-CAD exchange rate tracks the exchange rate between the U.S. and Canadian dollars affecting trade and investment.
    \item \texttt{exogenous\_fppi\_total}: Farm product price index tracks price changes for Canadian agricultural commodities reflecting market conditions.
\end{itemize}

\noindent The historical time series for the \texttt{target\_meat} variable is shown below:

\begin{verbatim}
<history>
REF_DATE,exogenous_fmpi,exogenous_commodity_food,exogenous_excaus,exogenous_fppi_total,target_meat
1986-01-01,99.8,46.12,1.407,77.5,65.1
1986-02-01,99.2,44.79,1.4043,77.4,64.2
1986-03-01,99.0,45.4,1.4009,76.4,64.2
1986-04-01,98.6,45.51,1.3879,75.3,63.6
1986-05-01,99.3,43.92,1.3757,75.9,64.0
1986-06-01,99.6,42.61,1.3899,76.8,64.9
1986-07-01,101.0,40.65,1.3808,78.4,66.5
1986-08-01,101.7,39.9,1.3885,76.0,67.8
1986-09-01,101.5,39.65,1.3873,76.0,71.3
1986-10-01,101.3,40.7,1.3885,76.6,71.5
1986-11-01,101.5,41.23,1.3863,77.1,72.1
1986-12-01,101.4,40.52,1.3801,75.7,72.9
% Add more rows as necessary...
</history>
\end{verbatim}

\noindent Please predict the \texttt{target\_meat} value at the REF\_DATEs below:

\begin{verbatim}
<target_dates>
2023-07-01
2023-08-01
2023-09-01
2023-10-01
2023-11-01
2023-12-01
2024-01-01
2024-02-01
2024-03-01
2024-04-01
2024-05-01
2024-06-01
</target_dates>
\end{verbatim}

\noindent Return the forecast in (REF\_DATE, \texttt{target\_meat}) format between \texttt{<forecast>} and \texttt{</forecast>} tags. Do not include any other information (e.g., comments) in the forecast.

\noindent Example format:
\begin{verbatim}
<forecast>
2023-07-01, v1
2023-08-01, v2
2023-09-01, v3
2023-10-01, v4
2023-11-01, v5
2023-12-01, v6
2024-01-01, v7
2024-02-01, v8
2024-03-01, v9
2024-04-01, v10
2024-05-01, v11
2024-06-01, v12
</forecast>
\end{verbatim}